\documentclass[sigconf,natbib=true,anonymous=false]{acmart}
\usepackage{booktabs}
\usepackage{multirow}
\usepackage{tabularx}
\usepackage{array}
\usepackage{float}
\usepackage{subcaption}
\usepackage[table]{xcolor}
\definecolor{lightpurple}{RGB}{235,230,250}
\newcolumntype{Y}{>{\centering\arraybackslash}X}

\AtBeginDocument{%
  }

\setcopyright{acmlicensed}
\copyrightyear{2026}
\acmYear{2026}
\acmDOI{XXXXXXX.XXXXXXX}
\acmConference[ SIGIR '26]{Make sure to enter the correct
  conference title from your rights confirmation email}{June 03--05,
  2026}{Melbourne, Australia}
\acmISBN{XXX-X-XXXX-XXXX-X/2026/06}
\begin{document}

\title{Multi-View Hierarchical Graph Neural Network for Sketch-Based 3D Shape Retrieval}

\author{Hang Cheng}
\authornote{Both authors contributed equally to this research.}
\affiliation{%
  \institution{Tsinghua Shenzhen International Graduate School, Tsinghua University}
  \city{Shenzhen}
  \state{Guangdong}
  \country{China}
}
\email{chenghan24@mails.tsinghua.edu.cn}

\author{Muyan He}
\authornotemark[1]
\affiliation{%
  \institution{Control and Simulation Center, Harbin Institute of Technology}
  \city{Harbin}
  \country{Heilongjiang}}
\email{24S104214@stu.hit.edu.cn}

\author{Mingyu Fan}
\affiliation{%
  \institution{Tsinghua Shenzhen International Graduate School, Tsinghua University}
  \city{Shenzhen}
  \state{Guangdong}
  \country{China}}
\email{my-fan25@mails.tsinghua.edu.cn}

\author{Chengfeng Xie}
\affiliation{%
 \institution{Tsinghua Shenzhen International Graduate School, Tsinghua University}
 \city{Shenzhen}
 \state{Guangdong}
  \country{China}}
\email{xiechenfeng@stu.hit.edu.cn}

\author{Xi Cheng}
\affiliation{%
  \institution{Tsinghua Shenzhen International Graduate School, Tsinghua University}
  \city{Shenzhen}
  \state{Guangdong}
  \country{China}}
\email{chengxi23@tsinghua.org.cn}

\author{Long Zeng}
\authornote{Corresponding author.}
\affiliation{%
  \institution{Tsinghua Shenzhen International Graduate School, Tsinghua University}
  \city{Shenzhen}
  \state{Guangdong}
  \country{China}}
\email{zenglong@sz.tsinghua.edu.cn}


\begin{abstract}
Sketch-based 3D shape retrieval (SBSR) aims to retrieve 3D shapes that are consistent with the category of the input hand-drawn sketch. The core challenge of this task lies in two aspects: existing methods typically employ simplified aggregation strategies for independently encoded 3D multi-view features, which ignore the geometric relationships between views and multi-level details, resulting in weak 3D representation. Simultaneously, traditional SBSR methods are constrained by visible category limitations, leading to poor performance in zero-shot scenarios. To address these challenges, we propose \textbf{Multi-View Hierarchical Graph Neural Network} (MV-HGNN), a novel framework for SBSR. Specifically, we construct a view-level graph and capture adjacent geometric dependencies and cross-view message passing via local graph convolution and global attention. A view selector is further introduced to perform hierarchical graph coarsening, enabling a progressively larger receptive field for graph convolution and mitigating the interference of redundant views, which leads to more discriminate discriminative hierarchical 3D representation. To enable category-agnostic alignment and mitigate overfitting to seen classes, we leverage CLIP text embeddings as semantic prototypes and project both sketch and 3D features into a shared semantic space. We use a two-stage training strategy for category-level retrieval and a one-stage strategy for zero-shot retrieval under the same model architecture. Under both category-level and zero-shot settings, extensive experiments on two public benchmarks demonstrate that MV-HGNN outperforms state-of-the-art methods.

\end{abstract}



\maketitle

\section{Introduction}
3D shape retrieval has become increasingly important with the rapid growth of 3D assets in gaming, virtual reality, digital twins, and robotics \cite{Xie2016TPAMI}. Traditional 3D retrieval methods based on text or images are often limited by ambiguous queries, domain gaps, and reliance on manual annotation, making them less effective for large-scale retrieval. Among cross-modal retrieval paradigms, hand-drawn sketches provide a particularly compelling interface: they are lightweight \cite{Bandyopadhyay2024CVPRExplain, Bandyopadhyay2024CVPRSketchINR, zeng2019sketch}, intuitive, and robust to texture interference, while effectively conveying object structure through sparse strokes. These properties make Sketch-Based 3D Shape Retrieval (SBSR) a promising tool for interactive design \cite{zeng2014sketch2jewelry, liao2024freehand}, creative modeling, and visual content generation \cite{Wang2015CVPR, Dai2017AAAI, Dai2018TIP}.

Existing literature generally attributes the primary difficulty in SBSR to the cross-modal domain gap: sketches are sparse 2D images, whereas 3D shapes possess highly complex geometric and topological structures \cite{Qi2018BMVC, Liang2021TIP, Dai2020ICME}. This asymmetry in representation renders the sketch-3D embedding spaces intrinsically inconsistent. To solve this problem, prior works have attempted to bridge this gap via Siamese networks \cite{Wang2015CVPR}, metric learning \cite{Dai2018TIP}, adversarial alignment\cite{Chen2018ECCV}, and triplet/center losses \cite{Qi2018BMVC}. Despite promising results, we observe that even under well-structured alignment strategies, the feature representation of 3D shapes remains an importance bottleneck affecting SBSR performance due to their complex geometric structures. Specifically, 3D shape data is high-dimensional, typically represented by voxel grids \cite{maturana2015voxnet}, polygon meshes, or point clouds\cite{Wang2019TOG, xu2018spidercnn}. Direct 3D representations are often computationally expensive and require substantial memory and processing overhead. In contrast, view-based approaches are generally more efficient and easier to deploy, which has motivated a shift toward this paradigm in recent years. These methods typically employ a shared feature extractor to process multi-view images rendered from 3D shapes, and aggregate the resulting view embeddings into a unified shape representation via pooling operations \cite{su2015multi}. However, simple pooling schemes tend to overlook fine-grained details within individual views and fail to explicitly capture the inherent relationships among different viewpoints \cite{Yang2022MMS, Bai2023KBS, Liang2024CVIU, bai2025scdl}. Although a few studies have explored attention-based fusion \cite{zhu2024sketch, zhang2025meha}, they often suffer from insufficient feature expression capabilities as they fail to explicitly model pose variations and correspondences between different viewpoints. Consequently, effectively capturing spatial and pose relationships among multiple viewpoints to model highly discriminative 3D representations with geometric properties remains a challenging problem.

On the other hand, most existing methods fail to address zero-shot scenarios and lack a unified framework adaptable to diverse settings. While prior works have achieved promising performance on seen categories, they heavily rely on embeddings tailored for closed-set classification, which substantially limits their generalization to unseen categories. A few methods dedicated to zero-shot learning \cite{Wang2023AAAI, Xu2022AAAI, Li2025TCSVT} learn features that lack informativeness and generalization, hampering the effective identification of unseen samples. Moreover, existing approaches are almost exclusively limited to independent models designed solely for category-level retrieval or zero-shot retrieval, lacking the universality to adapt to scenarios with varying data visibility. Therefore, constructing a generalizable cross-modal alignment space that extends to retrieval scenarios with unseen categories remains an under-resolved issue.

\begin{figure}
  \centering
  \includegraphics[
    width=\linewidth,
    trim=0mm 0mm 0mm 25mm,
    clip
]{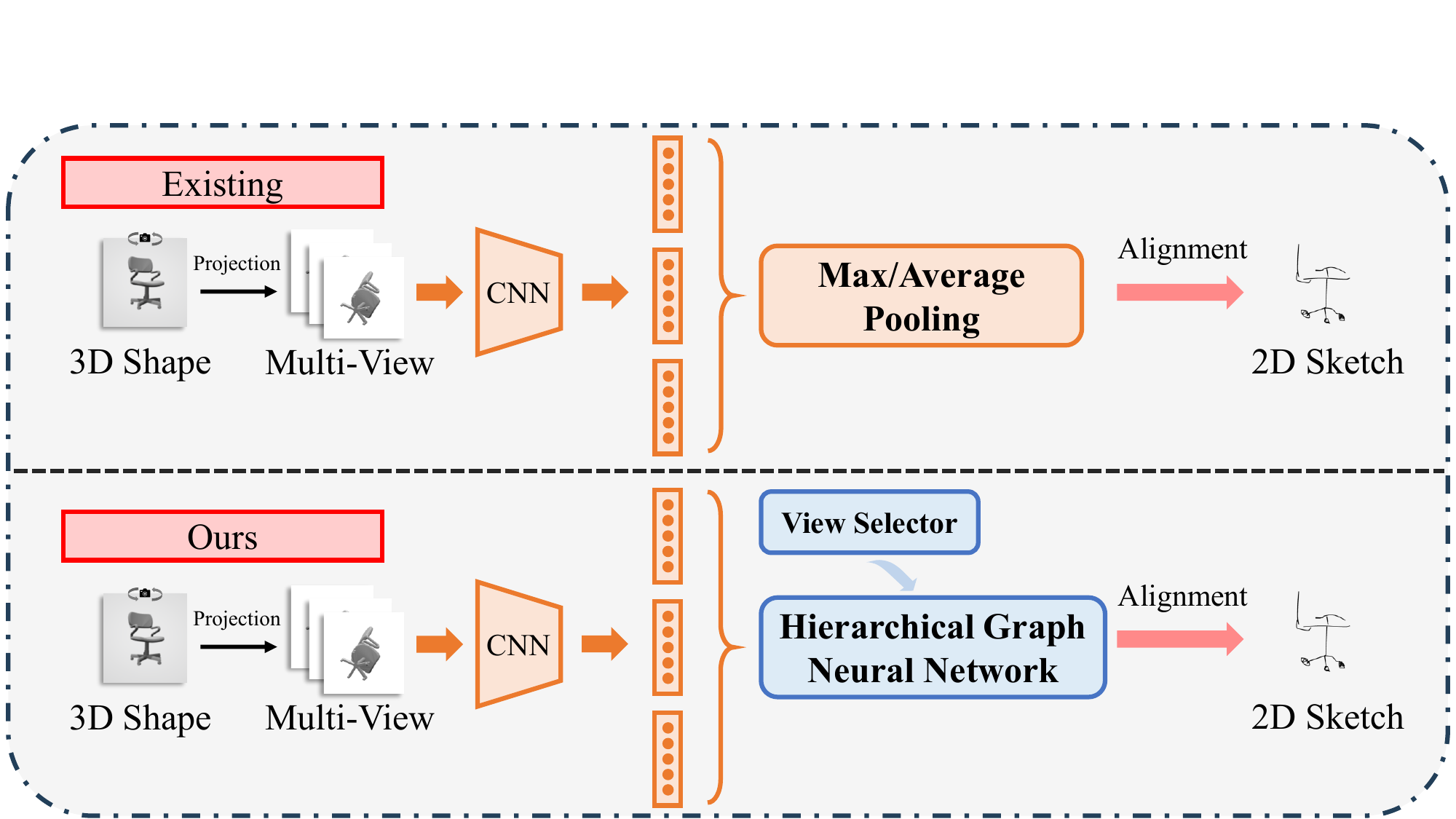}
  \caption{Existing method and our method.}
  \label{fig:intro}
\end{figure}

To address the two aforementioned problems, we propose MV-HGNN, a novel Hierarchical Graph Neural Network for SBSR. Specifically, for 3D shape feature extraction, we apply single-view encoding to different views of the 3D shape and construct a multi-view graph structure where viewpoints serve as nodes. We employ local geometry-aware graph convolutions to capture structural dependencies of adjacent views and global attention to model long-range relationships across views. Building on this observation, we introduce a learnable view selector that adaptively mitigates the interference of redundant views and progressively enlarges the effective receptive field of graph convolution. At each stage, the selected views are used to update the graph nodes and construct a coarser graph for the next level of reasoning(Fig. \ref{fig:intro}). By repeating this process for multiple scales, we obtain hierarchical features that encode geometric structure and yield a robust multi-level 3D shape representation. For the sketch modality, we naturally incorporate a visual prompt tuning\cite{Radford2021ICML} encoder for feature extraction.

Furthermore, we design a prototype learning strategy to establish an effective, aligned cross-modal embedding space to bridge the modal gap. Specifically, we leverage the CLIP \cite{Radford2021ICML} text space to construct category semantic prototypes, mapping sketch and 3D features into a unified prototype space, which ensures that the alignment strategy is built upon high-quality 3D representations. Simultaneously, for category-level SBSR, we employ a two-stage training strategy supervised by category labels: pre-training the 3D modality followed by the sketch modality. For zero-shot SBSR tasks, we adopt a joint training strategy for both modalities. This strategy enables the transition from designing for individual tasks to a more generalized retrieval framework managing various specialized scenarios. The main contributions of this paper are as follows:

\begin{itemize}
\item {We propose a novel Multi-View Hierarchical Graph Neural Network (MV-HGNN), addressing the challenges of low discriminative 3D shape representations and the non-universality under the zero-shot setting.
}
\item {We use multiple views to construct graph nodes and designing a hierarchical graph coarsening mechanism with a learnable view selector to progressively aggregate multi-view information and construct multi-level discriminative representations with geometric information.
}
\item {We introduce a CLIP-assisted prototype learning strategy that aligns sketches and 3D shapes, mitigating category bias and facilitating zero-shot generalization in SBSR, with a single model supporting both zero-shot and category-level retrieval via one-stage and two-stage training.
}
\item {Extensive experiments on two public benchmarks SHREC2013 and SHREC2014 demonstrate that our method outperforms existing state-of-the-art approaches in both category-level and zero-shot settings.
}
\end{itemize}

\section{Related Work}
\noindent\textbf{Sketch-based 3D Shape Retrieval.} Sketch-based 3D shape retrieval (SBSR) is a challenging cross-modal task. Early approaches relied on hand-crafted sketch features and shallow models \cite{Yoon2010ACMMM, Saavedra2012EG3DOR, Li2013SHREC, Li2014CVIU, Li2014SHREC}. With the advent of deep learning, subsequent methods \cite{Wang2015CVPR,Dai2018TIP,Chen2018ECCV,Qin2022CAG} achieved superior performance by learning nonlinear representations and mapping sketches and 3D shapes into a shared embedding space via metric learning \cite{Lei2019PR, Dai2017AAAI, Dai2018TIP}. To mitigate cross-modal discrepancies, adversarial learning has also been explored. Given the ill-posed nature of aligning sparse 2D sketches with complex 3D shapes, prior work has further investigated view-based modeling, attention mechanisms, and cross-modal feature interaction \cite{bai2023pagml, xu2020tmm, zhou2020tvcg, Zhao2022JVCIR}. Cross-modal view attention \cite{Qi2021TIP} and teacher–student alignment strategies \cite{Dai2020ICME,Liang2024CVIU} have been proposed to bridge the modality gap. Additionally, uncertainty modeling has been introduced to handle noisy and subjective sketches \cite{Liang2021TIP, Cai2023ICME}, while recent work leverages non-differentiable rasterization to model sketch stroke attributes \cite{Bai2023KBS,su2025dkd,su2025skd}. Despite these advances, most existing methods emphasize cross-modal alignment under seen-category settings and treat 3D representations as either fixed or weakly structured, limiting their effectiveness in zero-shot SBSR.

\noindent\textbf{3D Shape Learning for Sketch-based 3D Shape Retrieval.} Mapping 2D sketches to 3D shapes in SBSR is inherently ill-posed, as a single 3D object may correspond to multiple sketches from different viewpoints. Early model-based approaches represent 3D shapes using voxel grids \cite{maturana2015voxnet}, point clouds \cite{klokov2017escape, Qi2018BMVC}, or dedicated 3D encoders such as 3D CNNs \cite{wang2017cnn, esteves2018learning} and PointNet variants \cite{qi2017pointnet, qi2017pointnet++}, and further address pose correspondence via BPnP-based 2D–3D alignment \cite{chen2020end, Chowdhury2023ICCV}. However, their high computational cost limits scalability \cite{chen2025training}. View-based methods alleviate this issue by representing 3D shapes through projected 2D views. Early works learn intra- and cross-domain similarities from paired views \cite{Wang2015CVPR, yu2016sketch} or compute global descriptors using LD-SHIFT \cite{darom2012scale} and 3D-SHIFT with LLC \cite{Dai2017AAAI}. Most recent methods employ a shared CNN followed by global or max pooling \cite{su2015multi, Dai2020ICME, Cai2023ICME, Liang2024CVIU}, which is efficient but coarse and fails to explicitly capture inter-view relationships. More recent efforts adopt Transformer-based aggregation \cite{zhu2024sketch, zhang2025meha} or viewpoint selection strategies \cite{yuan2023retrieval}, yet they still largely rely on global aggregation that can obscure local geometric structures and fine-grained cross-view variations. We therefore contend that the quality of 3D representation learning remains a primary bottleneck in SBSR, motivating our structured, hierarchical representation that explicitly models inter-view relationships to better support cross-modal alignment.

\noindent\textbf{Zero-shot Retrieval.} Zero-shot retrieval aims to generalize to unseen categories at test time, posing substantial challenges for cross-modal generalization. Most existing zero-shot SBSR methods rely on semantic embeddings, attribute supervision, or pretrained vision–language models. In contrast, zero-shot sketch-based image retrieval has been more extensively studied, with approaches based on conditional generation \cite{yelamarthi2018zero}, cross-modal reconstruction \cite{deng2020progressive}, knowledge distillation \cite{tian2021relationship}, and test-time adaptation \cite{sain2022sketch3t}, which is further improved when combined with CLIP \cite{sain2023clip, lin2023zero, li2024dr, singha2024elevating}. By comparison, zero-shot SBSR remains underexplored. \cite{Xu2022AAAI} first extended SBSR to the zero-shot setting via adversarial learning with disentangled representations; \cite{Wang2023AAAI} incorporated prototype learning and stroke-level sketch representations; \cite{Chowdhury2023ICCV} leveraged images as an intermediate modality to learn 3D viewpoint knowledge; and \cite{Meng2024TCSVT} further enhanced semantic prototype learning. Prior work is task-specific to either category-level or zero-shot retrieval, limiting generalization across settings. We instead unify both within a single framework by modeling their distinct supervision characteristics.

\section{Method}
\subsection{Problem Definition.} Let $\mathcal{S}$ and $\mathcal{X}$ denote the sketch and 3D shape domains. Given a sketch encoder $F_{\text{ske}}(\cdot)$ and a 3D shape encoder $F_{\text{3D}}(\cdot)$, each sketch $s \in \mathcal{S}$ and 3D shape $x \in \mathcal{X}$ are mapped to features $F_{\text{ske}}(s)$ and $F_{\text{3D}}(x)$ in a shared embedding space. At inference, given a query sketch $s$, the system retrieves 3D shapes from a database $\mathcal{D}=\{x_i\}_{i=1}^N$ by ranking them according to a similarity measure between $F{_\text{ske}}(s)$ and ${F_{\text{3D}}(x_i)}_{i=1}^N$. A retrieval is deemed correct if the top-ranked shape shares the same semantic category as the query sketch$s$.

\subsection{Framework Overview.} 
\begin{figure*}[t]
  \centering
  \includegraphics[
    width=\textwidth,
    trim=0mm 5mm 0mm 3mm,
    clip
  ]{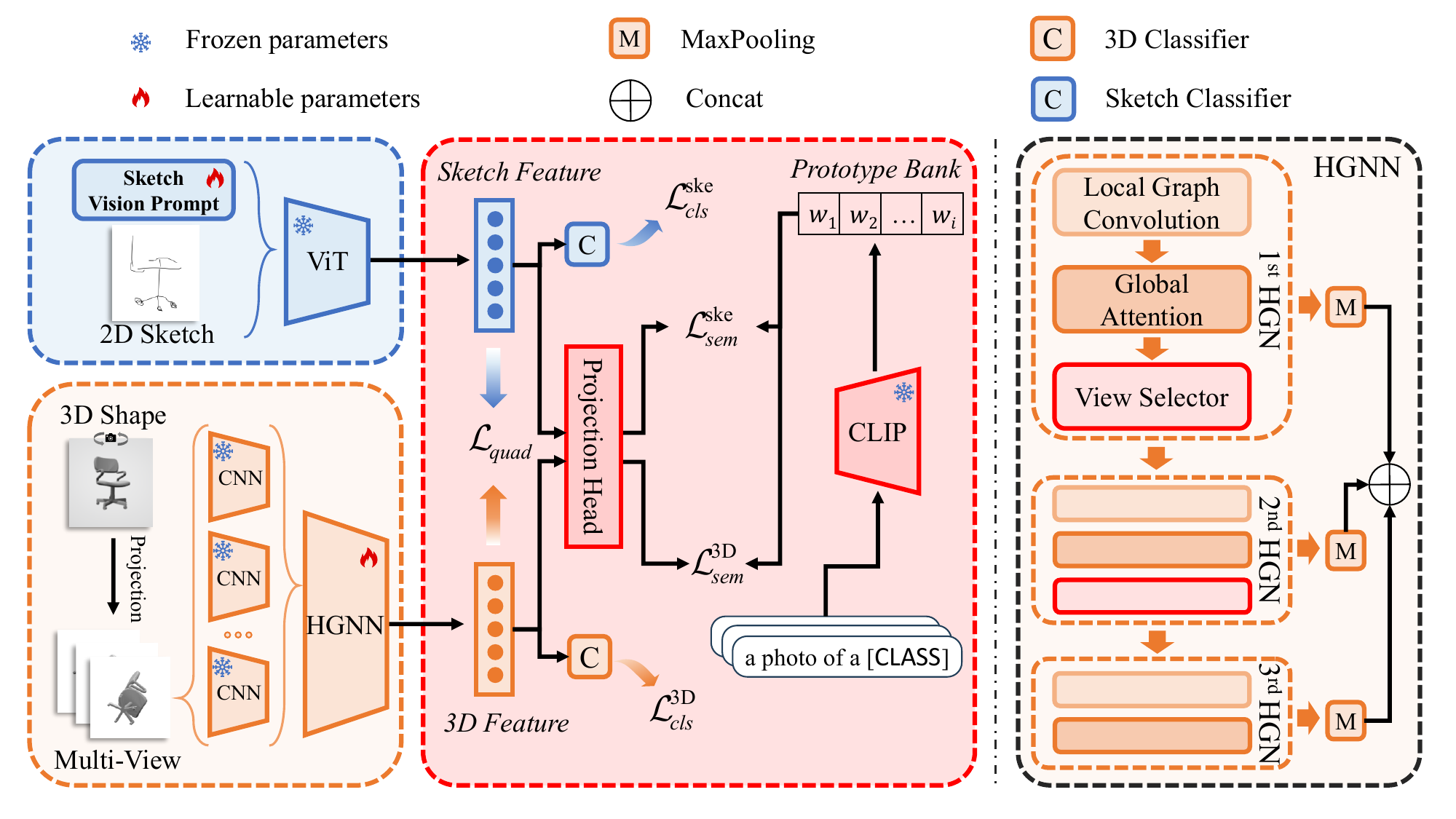}
  \caption{Overview of the proposed MV-HGNN framework. Left: Cross-modal alignment for 3D shape retrieval based on sketches. Sketch features and the overall representation of 3D shapes are aligned in a shared semantic space guided by prototype and assisted by CLIP text to support category-level and zero-shot retrieval. Right: Architecture of the HGNN 3D shape encoder. Each 3D shape is rendered into multiple views and encoded by a shared CNN. View features are organized into a geometrically-aware graph according to viewpoints, capturing neighborhood consistency and long-distance viewpoint dependence simultaneously through local graph convolution and global attention message passing. New viewpoints are updated through a view selector and the local graph convolution and global message passing are repeated, ultimately obtaining a multi-level 3D shape representation.}
  \label{fig:framework}
\end{figure*}

This work proposes a hierarchical graph neural network for view-based learning in sketch-based 3D shape retrieval. The framework comprises a 3D view feature extraction branch, a sketch feature extraction branch, a dual-branch discriminative classifier, and multiple cross-modal alignment losses. Its objective is to embed sketches and 3D shapes into a unified discriminative semantic space for cross-modal retrieval. An overview of the 3D shape representation framework is presented in Fig.~\ref{fig:framework}. During inference, the dual-branch classifier and semantic projection layers are discarded. Retrieval is performed by independently extracting features from 3D shapes and sketches using their respective encoders.

\subsection{3D Shape Encoder}
\subsubsection{Initial View Feature Extraction}

Given a 3D shape, we render a fixed number of views from predefined camera positions.
Let $\{I_i\}_{i=1}^{V}$ denote the rendered views, where $V$ is the number of viewpoints (typically $V=12$). Each view is processed by a shared CNN backbone (ResNet-18 pretrained on ImageNet), producing view-level feature vectors:
\begin{equation}
\mathbf{F}^{(0)} = \{ \mathbf{f}_i^{(0)} \}_{i=1}^{V}, \quad \mathbf{f}_i^{(0)} \in \mathbb{R}^{d},
\end{equation}
where $d = 512$. The backbone parameters are frozen during training. 

We associate each view with its 3D camera position $\mathbf{v}_i \in \mathbb{R}^3$, forming an initial node set $\mathcal{V}^{(0)} = \{\mathbf{v}_i\}_{i=1}^V$. We construct a directed view graph $G = (\mathcal{V}, \mathcal{E})$, where
\begin{equation}
\mathcal{V} = \{1, 2, \dots, V\},
\end{equation}
and the edge set $\mathcal{E}$ is defined as
\begin{equation}
\mathcal{E} = \{ (i,j) \mid \mathbf{v}_j \in \mathrm{kNN}(\mathbf{v}_i) \}.
\end{equation}
where $\mathrm{kNN}(\mathbf{v}_i)$ denotes the set of $k$ nearest camera positions to $\mathbf{v}_i$ measured by Euclidean distance.

\subsubsection{Local Graph Convolution}
Given the view graph $G^{(l)}$ at layer $l$, which consists of $N_l$ nodes (views) with node features stored as rows of $\mathbf{F}^{(l)}$, we define a local graph convolution layer to update node features by modeling relationships among neighboring views determined by $k$-nearest neighbors in the camera coordinate space. 

Local graph convolution enforces geometric consistency among neighboring views.
For each node $i$, we identify its $k$ nearest neighbors $\mathcal{N}(i)$ based on Euclidean distance between camera positions.

Attention weights are computed using scaled dot-product similarity:
\begin{equation}
\mathrm{sim}_{ij} = \frac{\mathbf{f}_i^\top \mathbf{f}_j}{\sqrt{d}}, \quad
\alpha_{ij} =
\frac{\exp(\mathrm{sim}_{ij})}
{\sum_{j' \in \mathcal{N}(i)} \exp(\mathrm{sim}_{ij'})}.
\end{equation}

Given the node feature matrix $\mathbf{F}^{(l)} \in \mathbb{R}^{N_l \times d}$ at layer $l$, the local graph convolution is defined as
\begin{equation}
\mathbf{F}^{(l+1)}
= \Psi\!\left(
\mathbf{A}^{(l)} \mathbf{F}^{(l)} \mathbf{W}^{(l)}
\right),
\end{equation}
where $\mathbf{A}^{(l)}$ is a row-normalized attention-weighted adjacency matrix with self-loops,
$\mathbf{W}^{(l)} \in \mathbb{R}^{d \times d}$ is a learnable weight matrix,
and $\Psi(\cdot)$ denotes Batch Normalization followed by LeakyReLU.

This formulation integrates geometric neighborhood priors with feature-adaptive aggregation.

\subsubsection{Global Attention Message Passing}
To complement local graph convolution, we introduce a non-local attention module to capture long-range dependencies across all views. Given node features $\mathbf{F}^{(l)} \in \mathbb{R}^{N_l \times d}$, we first project them into query, key, and value spaces via $\mathbf{Q} = \mathbf{F}^{(l)} \mathbf{W}_q$, $\mathbf{K} = \mathbf{F}^{(l)} \mathbf{W}_k$, and $\mathbf{V} = \mathbf{F}^{(l)} \mathbf{W}_v$, where $\mathbf{W}_q, \mathbf{W}_k \in \mathbb{R}^{d \times d/2}$ and $\mathbf{W}_v \in \mathbb{R}^{d \times d}$. Non-local attention weights are computed as $\mathbf{A} = \mathrm{softmax}\big( \mathbf{Q}\mathbf{K}^\top / \sqrt{d/2} \big)$ over all view pairs, followed by aggregation $\tilde{\mathbf{F}}^{(l)} = \mathbf{A}\mathbf{V}$.  

\begin{equation}
\mathbf{F}^{(l)}_{\text{nl}} = \mathbf{F}^{(l)} + \tilde{\mathbf{F}}^{(l)},
\end{equation}
where a residual connection stabilizes training. 

\subsubsection{View Prototype Selector}
To reduce redundancy and build hierarchical representations, we introduce a learnable view prototype selector. Given a feature matrix $\mathbf{F} \!\in\! \mathbb{R}^{V \times d}$, attention weights are computed as $\mathbf{A} = \mathrm{softmax}(\mathrm{MLP}(\mathbf{F}^{T})) \in \mathbb{R}^{K \times V}$ along the view dimension, where $K$ is the number of prototypes. Prototype features are then obtained by weighted aggregation, $\mathbf{P} = \mathbf{A}\mathbf{F}$, and camera positions by $\mathbf{v}_p = \mathbf{A}\mathbf{v}$, enabling geometry-aware kNN construction. The soft View Selector is fully differentiable and end-to-end trainable. Unlike standard attention pooling, it generates multiple prototypes via independent soft assignments, each capturing distinct view-level patterns, preserving complementary geometric cues, and effectively enlarging the receptive field for local graph convolution.

\subsubsection{Hierarchical View-GCN Pipeline}

The complete View-GCN is constructed by stacking multiple stages, each consisting of:
\begin{equation}
\text{LocalGCN} \rightarrow \text{Global Attention} \rightarrow \text{ViewSelector}.
\end{equation}

Starting from the initial view graph, the number of nodes is progressively reduced (e.g., $12 \rightarrow 6 \rightarrow 3$), forming a hierarchical graph structure. At each level, max pooling is applied over node features to obtain a level-specific representation. Finally, pooled features from all levels are concatenated and passed through an MLP to produce the final 3D shape representation.

\subsection{Sketch Encoder}

Due to substantial style variations and noise in sketches, achieving reliable alignment between sketch and 3D modalities is non-trivial. To address this issue, we adopt a CLIP \cite{Radford2021ICML} image encoder with vision prompt tuning \cite{jia2022vpt} to extract initial sketch representations. Specifically, visual prompts are injected into the input to guide CLIP toward modeling the sketch distribution, formulated as
\begin{equation}
\mathbf{V} = (\mathbf{V}_s, \mathbf{V}_p), \quad \mathbf{V} \in \mathbb{R}^{N_C \times d}.
\end{equation}
During training, the knowledge learned by CLIP is distilled into the prompt parameters via backpropagation. Only the layer normalization parameters and the prompt embeddings within the transformer encoder are updated, while all other CLIP weights remain frozen. This design preserves the generalization capability of the pretrained model while allowing limited adaptation to the sketch modality. By restricting parameter updates to a small set of normalization and prompt parameters, the proposed strategy maintains training stability, mitigates catastrophic forgetting, and reduces the risk of overfitting during cross-modal semantic alignment.

\subsection{Category-Level Training Strategy}
\begin{figure}
  \centering
  \includegraphics[
    width=\linewidth,
    trim=0mm 0mm 0mm 55mm,
    clip
]{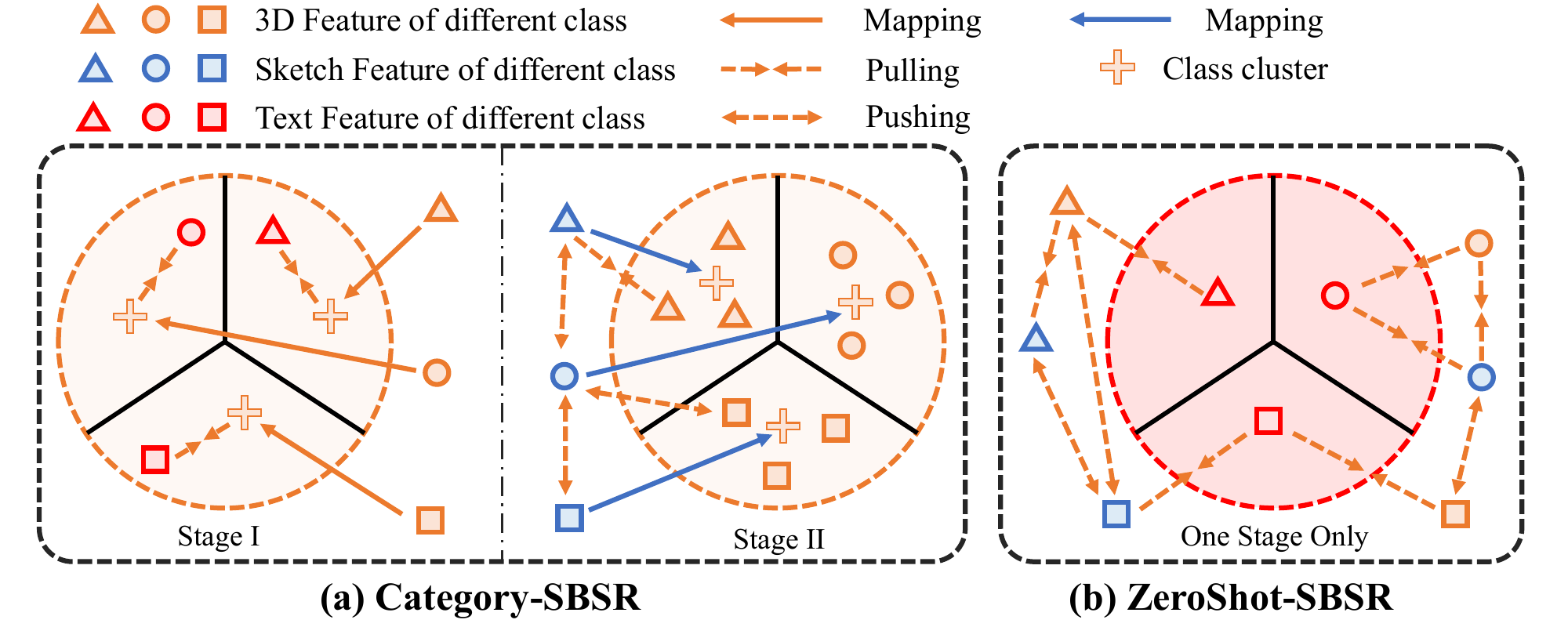}
  \caption{Illustration of the training strategies under category-level and zero-shot settings. Different shapes represent different object categories.}
  \label{fig:train}
\end{figure}

The geometric structure of 3D shapes is inherently complex, and the quality of their representations is fundamental to achieving high-performance SBSR. Introducing sketches for joint optimization too early may distract the model from effectively modeling 3D geometric information. Moreover, to mitigate the adverse impact of sketch noise on 3D shape representations, we adopt a two-stage strategy. Specifically, we first learn a robust 3D shape representation that fully captures its structural characteristics, and then perform cross-modal alignment to enable more effective sketch–shape matching and enhance overall model robustness (Fig.~\ref{fig:train}).

\subsubsection{Stage I: Pre-learned 3D}
In the first stage, the 3D shape encoder $F_{3D}$ is trained independently to construct a pretrained 3D shape prototype space enriched with semantic information. Category imbalance is a common issue in practical scenarios, which hinders the model from effectively learning the distributions of under-represented categories. To address this problem, we introduce semantic prototypes into the 3D shape learning process to provide a stable and category-agnostic metric that is particularly suitable for cross-modal alignment. 
Specifically, we construct a prototype bank defined as:

\begin{equation}
\mathcal{W} = \{\mathbf{w}_i\}_{i=1}^{C},
\end{equation}
where ${w}_i$ denotes the text embedding of the $i$-th category.
These embeddings are generated using prompt templates, where the class label replaces the \texttt{[CLASS]} token.

We employ a projection head $G(\cdot)$ to map 3D shape features ${f}_{3D}$ into the semantic prototype space, yielding ${p}_{3D}$, which is then aligned with the CLIP text embedding ${w}_y$.
The semantic alignment loss is formulated as:

\begin{equation}
\mathcal{L}^\mathrm{3D}_{sem} =
- \log
\frac{
\exp \left( \cos\left( \ell_2({p}_\mathrm{3D}), {w}_y \right) / \tau \right)
}{
\sum\limits_{{w}_i \in {W}}
\exp \left( \cos\left( \ell_2({p}_\mathrm{3D}), {w}_i \right) / \tau \right)
},
\end{equation}
where $\ell_2(\cdot)$ denotes $\ell_2$ normalization, $y$ is the ground-truth category label, and $\tau$ is a temperature parameter.

Semantic prototypes ${w}_c$ are constructed using the CLIP text encoder with prompt templates.
Classification is optimized using AM-Softmax. The loss function of the teacher classification network is defined as:

\begin{equation}
\mathcal{L}_{cls}^\mathrm{3D}
= -
\frac{1}{M}
\sum_{i=1}^{M}
\log
\frac{
\exp\Big\{{\, t\left(\bar{\mathbf{w}}_{y_i}^v \cdot \bar{{f}}_i^v - m\right)}\Big\}
}
{
\exp\Big\{\, t\left(\bar{\mathbf{w}}_{y_i}^v \cdot \bar{{f}}_i^v - m\right)\Big\}
+
\sum\limits_{j=1,\, j \neq y_i}^{N}
\exp\Big\{\, t\left(\bar{\mathbf{w}}_j^v \cdot \bar{{f}}_i^v\right)\Big\}
}
\label{eq:am_softmax}
\end{equation}

where $M$ is the mini-batch size and $y_i$ denotes the ground-truth label of the $i$-th sample. The feature vector ${f}_i^v$ represents the output of the 3D shape feature extractor for the $i$-th sample.

The normalized feature vector $\bar{{f}}_i^v$ and the normalized classifier weight $\bar{\mathbf{w}}_{y_i}^v$ are defined as:

\begin{equation}
\bar{{f}}_i^v = \frac{{f}_i^v}{\|{f}_i^v\|_2}, \quad
\bar{\mathbf{w}}_{y_i}^v = \frac{{\mathbf{w}}_{y_i}^v}{\|{\mathbf{w}}_{y_i}^v\|_2}.
\end{equation}

The scalar $t$ is a scale factor that controls the convergence behavior of the softmax function, while $m$ denotes the additive cosine margin that enforces a stricter decision boundary. This stage yields a discriminative and semantically structured 3D embedding space. The learning objective for the stage I can be denoted as:

\begin{equation}
\mathcal{L}^{\mathrm{SBSR}}_\mathrm{3D} =
\mathcal{L}_{cls}^\mathrm{3D}
+ \mathcal{L}_{sem}^\mathrm{3D}
\end{equation}

\subsubsection{Stage II: Cross-Modal Alignment}

The pretrained 3D encoder is frozen. The sketch encoder is trained to align with the learned 3D semantic space using a combination of losses, including classification loss on sketches, quadruplet loss for instance-level alignment, and semantic alignment loss to CLIP-based prototypes.

Due to the substantial modality gap, directly matching sketches to 3D shapes remains a challenging problem.
To address this issue, we align the sketch distribution with a pretrained 3D shape feature space.
To facilitate the mapping from sketches to the pretrained 3D feature space, sketches are classified using a frozen 3D shape classifier.

Furthermore, to ensure that sketches and 3D shapes from the same category are closer in the shared embedding space, we employ a quadruplet loss, defined as:
\begin{equation}
\begin{aligned}
\mathcal{L}_{quad} =
& \max\!\left(0, \mu + \delta({f}^{\mathrm{ske}}_a, {f}^{\mathrm{3D}}_p)
- \delta({f}^{\mathrm{ske}}_a, {f}^{3D}_n)\right) + \\
& \max\!\left(0, \mu + \delta({f}^{\mathrm{ske}}_a, {f}^{\mathrm{3D}}_p)
- \delta({f}^{\mathrm{ske}}_a, {f}^{\mathrm{ske}}_n)\right)
\end{aligned}
\end{equation}

where $\mu$ denotes the margin parameter and $\delta(\cdot,\cdot)$ represents the distance metric.
Given a quadruplet $({f}^{\mathrm{ske}}_a, {f}^{\mathrm{3D}}_p, {f}^{{3D}}_n, {f}^{\mathrm{ske}}_n)$, ${f}^{\mathrm{ske}}_a$ is the sketch anchor, ${f}^\mathrm{3D}_p$ is the corresponding 3D feature from the same category, and ${f}^{\mathrm{3D}}_n$ and ${f}^{\mathrm{ske}}_n$ denote 3D and sketch features from different categories, respectively.

The overall objective is
\begin{equation}
\mathcal{L}^{\mathrm{SBSR}}_{total} =
 \mathcal{L}_{quad}
+ \mathcal{L}^{\mathrm{ske}}_{cls}
+ \mathcal{L}^{\mathrm{ske}}_{sem}
\end{equation}

\subsection{Zero-Shot Training Strategy}

Unlike category-level SBSR, where two-stage training can first optimize a stable 3D representation and then align sketches to it, such strategy in the zero-shot setting risks overfitting the 3D encoder to seen categories. In contrast, zero-shot SBSR requires representations that generalize to unseen categories, where overly category-specific optimization can hinder semantic generalization. To this end, we adopt a unified one-stage joint training strategy in which sketch and 3D encoders are optimized simultaneously and projected into a shared semantic space defined by CLIP text prototypes. By enforcing prototype-based cross-modal alignment together with quadruplet constraints, the learned embedding space becomes more category-agnostic and naturally generalizes to unseen categories without relying on explicit class supervision. These observations highlight an inherent trade-off between discriminability on seen categories and generalization to unseen semantics (Fig.~\ref{fig:train}). The overall objective function for zero-shot SBSR is defined as:

\begin{equation}
\mathcal{L}^{\mathrm{ZS-SBSR}}_{total} =
\mathcal{L}_{quad}
+ \mathcal{L}^{\mathrm{3D}}_{sem}
+ \mathcal{L}^{\mathrm{ske}}_{sem}
\end{equation}

\section{Experiments}
\subsection{Experiment Setup}
\subsubsection{Datasets}

We evaluate our method on two standard benchmarks, SHREC’13~\cite{Li2013SHREC} and SHREC’14~\cite{Li2014SHREC}. SHREC’13 comprises 90 categories with 80 sketches and 1,258 3D shapes from PSB and hand-drawn sketches, while SHREC’14 contains 171 categories with 13,680 sketches and 8,987 3D shapes. Both datasets follow the standard 50--30 sketch split. For zero-shot evaluation, 23 (SHREC’13) and 38 (SHREC’14) categories with $\leq$5 3D shapes are treated as unseen, with the remaining 67 and 133 categories as seen, respectively. For category-level SBSR, seen classes are further split: sketches retain the 50--30 split, and 3D shapes are partitioned into 80\% training and 20\% testing sets~\cite{Wang2023AAAI}.

\subsubsection{Evaluation Metrics}
Following the evaluation protocol in \cite{Wang2023AAAI}, we adopt a comprehensive set of retrieval metrics to assess the performance of our method. Specifically, we report Nearest Neighbor (NN), First-Tier (FT), Second-Tier (ST), Normalized Discounted Cumulative Gain (nDCG), E-measure (E), Mean Reciprocal Rank (MRR), and Mean Average Precision (mAP). Among these metrics, E-measure (E) is a distance-based evaluation criterion, where a lower value indicates better performance. For all other metrics, higher values correspond to superior retrieval accuracy.

\subsubsection{Implementation Details}
All experiments are conducted on a single NVIDIA RTX 4090 GPU. Our method is implemented in PyTorch, with multi-view 3D renderings generated via Blender. Sketches and 2D views of 3D shapes are resized to $224 \!\times\! 224$. We employ a CLIP text encoder (ViT-B/16) and ResNet-18 and ViT-B/16 visual encoders, all initialized with pretrained weights from ImageNet or large-scale image–text datasets. Each mini-batch contains 512 quadruplets (anchor sketch, positive 3D shape, negative 3D shape, negative sketch) with batch size 32, trained for 100 epochs. Optimization uses Adam with an initial learning rate of $1\!\times\!10^{-4}$, decayed to $1\!\times\!10^{-6}$ via a cosine scheduler. Hyperparameters are set as follows: temperature $\tau = 0.07$, quadruplet margin $1.0$, and classification loss parameters $t = 15$, $m = 0.3$.

\subsection{Comparison to State-of-the-Art Methods}

\begin{table*}[t]
  \caption{Comparison with state-of-the-art methods on Category-level SBSR and Zero-shot SBSR for benchmark SHREC’2013.}
  \label{tab:main_shrec13}
  \centering
  \setlength{\tabcolsep}{2.5pt}
  \begin{tabularx}{\textwidth}{c c c *{7}{Y} *{7}{Y}}
    \toprule
    \multirow{2}{*}{Type} & \multirow{2}{*}{Method} & \multirow{2}{*}{Publication}
    & \multicolumn{7}{c}{Category SBSR}
    & \multicolumn{7}{c}{Zero-Shot SBSR} \\
    \cmidrule(lr){4-10} \cmidrule(lr){11-17}
    & & 
    & NN$\uparrow$ & FT$\uparrow$ & ST$\uparrow$ & nDCG$\uparrow$ & E$\downarrow$ & MRR$\uparrow$ & mAP$\uparrow$
    & NN$\uparrow$ & FT$\uparrow$ & ST$\uparrow$ & nDCG$\uparrow$ & E$\downarrow$ & MRR$\uparrow$ & mAP$\uparrow$
    \\
    \midrule

    \multirow{3}{*}{3D Enc}
    & SDFSketch \cite{park2019deepsdf} & CVPR2019
    & 23.5 & 16.7 & 15.5 & 23.0 & 77.2 & 37.9 & 23.6
    & 15.7 & 13.8 & 11.7 & 18.6 & 75.2 & 25.3 & 22.3 \\
    & FGPointNet \cite{qi2017pointnet++} & NIPS2017
    & 46.7 & 48.7 & 35.7 & 59.1 & 52.6 & 53.2 & 57.3
    & 32.0 & 27.9 & 25.5 & 36.2 & 64.0 & 46.0 & 36.1 \\
    & FGSpherical \cite{esteves2018learning} & ECCV2018
    & 45.3 & 46.9 & 34.3 & 56.9 & 54.6 & 51.7 & 56.9
    & 29.2 & 19.7 & 23.1 & 33.4 & 66.1 & 43.9 & 33.5 \\
    \midrule

    \multirow{7}{*}{View Enc}
    & Siamese \cite{Wang2015CVPR} & CVPR2015
    & 19.0 & 14.2 & 12.5 & 20.8 & 76.1 & 33.2 & 20.8
    & 13.7 & 12.4 & 11.7 & 18.6 & 76.2 & 25.9 & 20.1 \\
    & DCHML \cite{Dai2018TIP} & TIP2018
    & 41.7 & 43.3 & 29.5 & 51.0 & 54.7 & 48.9 & 52.5
    & 18.3 & 21.1 & 13.6 & 23.8 & 63.2 & 27.8 & 33.9 \\
    & TCL \cite{he2018triplet} & CVPR2018
    & 43.3 & 47.7 & 33.7 & 57.2 & 51.5 & 51.9 & 56.6
    & 25.0 & 17.8 & 20.3 & 31.4 & 65.0 & 39.0 & 31.7 \\
    & CGN \cite{Dai2020ICME} & ICME2020
    & 65.0 & 52.0 & 30.8 & 60.8 & 53.4 & 77.5 & 58.8
    & 33.3 & 29.4 & 22.5 & 39.8 & 63.4 & 48.4 & 38.6 \\
    & PCL \cite{Wang2023AAAI} & AAAI2023
    & 55.0 & 54.7 & \underline{41.7} & \underline{71.9} & 45.4 & 68.7 & 65.9
    & \underline{38.3} & \textbf{38.9} & \underline{26.1} & \underline{47.0} & \textbf{54.9} & \underline{52.4} & \underline{48.0} \\
    & CFTTSL \cite{Liang2024CVIU} & CVIU2024
    & 75.3 & \underline{70.2} & 39.7 & 71.7 & \underline{39.5} & 80.8 & 74.8
    & 34.9 & 26.1 & 19.1 & 28.1 & 70.4 & 46.6 & 32.5 \\
    & MEHA \cite{zhang2025meha} & SIGIR2025
    & \underline{79.0} & 68.6 & {40.6} & 68.9 & {45.4} & \underline{83.3} & \underline{75.2}
    & 33.9 & 27.7 & {22.3} & 46.9 & 70.5 & 47.8 & 34.9 \\
    \midrule

    \rowcolor{lightpurple}
    & \textbf{MV-HGNN(Ours)} &  
    & \textbf{90.6} & \textbf{76.5} & \textbf{43.2} & \textbf{80.3} & \textbf{35.6} & \textbf{94.3} & \textbf{82.2}
    & \textbf{42.2} & \underline{34.3} & \textbf{28.2} & \textbf{52.9} & \underline{62.5} & \textbf{56.3} & \textbf{48.5} \\
    \bottomrule
  \end{tabularx}
\end{table*}

\begin{table*}[t]
  \caption{Comparison with state-of-the-art methods on Category-level SBSR and Zero-shot SBSR for benchmark SHREC’2014.}
  \label{tab:main_shrec14}
  \centering
  \setlength{\tabcolsep}{2.5pt}
  \begin{tabularx}{\textwidth}{c c c *{7}{Y} *{7}{Y}}
    \toprule
    \multirow{2}{*}{Type} & \multirow{2}{*}{Method} & \multirow{2}{*}{Publication}
    & \multicolumn{7}{c}{Category SBSR}
    & \multicolumn{7}{c}{Zero-Shot SBSR} \\
    \cmidrule(lr){4-10} \cmidrule(lr){11-17}
    & & 
    & NN$\uparrow$ & FT$\uparrow$ & ST$\uparrow$ & nDCG$\uparrow$ & E$\downarrow$ & MRR$\uparrow$ & mAP$\uparrow$
    & NN$\uparrow$ & FT$\uparrow$ & ST$\uparrow$ & nDCG$\uparrow$ & E$\downarrow$ & MRR$\uparrow$ & mAP$\uparrow$
    \\
    \midrule

    \multirow{3}{*}{3D Enc}
    & SDFSketch \cite{park2019deepsdf} & CVPR2019
    & 21.3 & 14.9 & 20.6 & 26.7 & 72.1 & 35.4 & 22.3
    & 14.1 & 12.5 & 10.8 & 15.9 & 76.8 & 24.5 & 20.4 \\
    & FGPointNet \cite{qi2017pointnet++} & NIPS2017
    & 39.5 & 41.6 & 23.4 & 39.1 & 62.3 & 46.1 & 42.3
    & 25.4 & 23.2 & 19.0 & 33.8 & 62.1 & 44.3 & 34.9 \\
    & FGSpherical \cite{esteves2018learning} & ECCV2018
    & 36.7 & 35.6 & 21.7 & 38.6 & 64.6 & 43.5 & 36.9
    & 22.9 & 22.5 & 18.3 & 32.8 & 65.7 & 42.3 & 32.7 \\
    \midrule

    \multirow{7}{*}{View Enc}
    & Siamese \cite{Wang2015CVPR} & CVPR2015
    & 18.7 & 12.9 & 12.0 & 20.0 & 76.1 & 31.2 & 20.7
    & 10.0 & 8.6 & 8.7 & 13.6 & 78.4 & 21.4 & 17.0 \\
    & DCHML \cite{Dai2018TIP} & TIP2018
    & 30.7 & 32.5 & 24.6 & 41.4 & 61.2 & 39.7 & 41.9
    & 16.7 & 17.7 & 17.9 & 27.3 & 67.3 & 27.8 & 30.0 \\
    & TCL \cite{he2018triplet} & CVPR2018
    & 33.3 & 31.7 & 20.9 & 38.0 & 63.6 & 40.7 & 40.4
    & 21.3 & 21.6 & 19.1 & 30.6 & 66.5 & 31.4 & 32.8 \\
    & CGN \cite{Dai2020ICME} & ICME2020
    & 42.0 & 35.0 & 22.2 & 41.7 & 61.6 & 46.8 & 44.2
    & 26.3 & 25.0 & 21.4 & 35.1 & 65.0 & 36.8 & 35.6 \\
    & PCL \cite{Wang2023AAAI} & AAAI2023
    & 43.0 & 40.3 & 24.3 & 45.5 & 58.1 & 48.2 & 49.1
    & \underline{33.3} & \underline{32.8} & \underline{22.8} & \underline{39.9} & \underline{57.8} & \underline{45.5} & \underline{43.4} \\
    & CFTTSL \cite{Liang2024CVIU} & CVIU2024
    & \underline{61.3} & \underline{56.3} & \underline{32.4} & \underline{62.8} & \underline{53.1} & \underline{69.3} & \underline{58.1}
    & 16.7 & 14.4 & 11.3 & 35.8 & 76.7 & 26.6 & 21.6 \\
    & MEHA \cite{zhang2025meha} & SIGIR2025
    & 32.8 & 24.2 & 34.5 & 44.4 & 75.5 & 42.8 & 26.7
    & 10.8 & 10.4 & 15.8 & 30.7 & 89.4 & 20.1 & 15.3 \\
    \midrule

    \rowcolor{lightpurple}
    & \textbf{MV-HGNN(Ours)} &  
    & \textbf{82.3} & \textbf{63.5} & \textbf{35.9} & \textbf{69.6} & \textbf{46.8} & \textbf{87.2} & \textbf{66.0}
    & \textbf{36.7} & \textbf{31.4} & \textbf{23.8} & \textbf{41.7} & \textbf{55.3} & \textbf{49.2} & \textbf{43.8} \\
    \bottomrule
  \end{tabularx}
\end{table*}

We compare our method with several state-of-the-art 2D view-based method, including Siamese \cite{Wang2015CVPR}, DCHML \cite{Dai2018TIP}, TCL \cite{he2018triplet}, CGN \cite{Dai2020ICME}, PCL \cite{Wang2023AAAI}, CFTTSL \cite{Liang2024CVIU}, and MEHA \cite{zhang2025meha}. Following the original papers, the Siamese method adopts a three-layer CNN backbone, AlexNet is used for DCHML, ResNet-50 is used for TCL and CGN, CFTTSL and MEHA employ the same pretrained ViT-B/16 backbone \cite{dosovitskiy2020vit}. CFTTSL and MEHA are specifically designed for the category-level SBSR task. We further compare with several model-based(3D Enc) approaches, including SDFSketch \cite{park2019deepsdf}, FGPointNet \cite{qi2017pointnet++}, FGSpherical \cite{esteves2018learning}. Since no official implementation is available for MEHA \cite{zhang2025meha}, we re-implement it following Section 4.2 of the original paper. We use the same backbone, optimizer, and training schedule as reported. The experimental results on the SHREC’13 and SHREC’14 datasets are reported in Tab.~\ref{tab:main_shrec13} and Tab.~\ref{tab:main_shrec14}. 

\noindent\textbf{For category-level SBSR}, MV-HGNN achieves substantial improvements over existing view-based approaches (e.g., PCL and MEHA), particularly on NN, nDCG, and mAP. Compared with MEHA, which also leverages multi-view representations, MV-HGNN yields notable gains (+11.6 in NN and +7.0 in mAP), suggesting that explicitly modeling inter-view geometric dependencies via a hierarchical graph leads to more discriminative 3D representations than simple view aggregation. Moreover, MV-HGNN achieves the lowest E-measure, indicating more reliable ranking quality across different recall levels.

\noindent\textbf{For zero-shot SBSR}, MV-HGNN also shows clear advantages over previous methods, achieving the best performance on most metrics (e.g., NN, ST, nDCG, MRR, and mAP). Notably, while some methods (e.g., CFTTSL) obtain competitive results on specific metrics, they suffer from performance degradation when generalizing to unseen categories. In contrast, MV-HGNN maintains stable retrieval performance, which validates the benefit of our one-stage joint training strategy and prototype-based alignment in learning more category-agnostic representations. Taken together, these results demonstrate that MV-HGNN not only enhances discriminability on seen categories but also improves generalization to unseen categories, highlighting a favorable trade-off between category-level accuracy and zero-shot transferability.

\subsection{Ablation Study}
\subsubsection{The Influence of Components of 3D Shape Encoder}
We conduct a comprehensive ablation study on SHREC2013 to quantify the contribution of encoder components (Table~\ref{tab:3d_Components}). Removing the local GCN or global attention reduces mAP from 82.2\% to 74.6\% and 73.3\%, respectively, highlighting their roles in preserving local geometry and modeling long-range cross-view dependencies. Disrupting view structure—via random order or random selector—lowers mAP to 72.5\% and 71.7\%, while replacing FPS with content-based selection also degrades performance, confirming that attention-guided, geometry-aware view selection is critical. For pooling, average pooling performs worst (mAP 73.7\%), attention pooling provides moderate gains (76.2\%), and ViT pooling improves further (79.8\%), yet hierarchical multi-level pooling combined with view selection achieves the best results (mAP 82.2\%, NN 90.6\%, nDCG 80.3\%), demonstrating that explicit view modeling and hierarchical compression outweigh pooling choice alone. Integrating first-, second-, and third-level features progressively improves metrics—first-level alone underperforms (mAP 70.1\%), first+second gains (76.7\%), and full hierarchy reaches the peak—validating our hierarchical representation design. Overall, metrics across NN, nDCG, and mAP consistently reflect the importance of structured view modeling, multi-level feature aggregation, and cross-view attention.

\begin{table}[t]
\centering
\caption{Ablation study of 3D shape encoder SHREC2013 Performance Comparison.}
\label{tab:3d_Components}
\begin{tabularx}{\linewidth}{>{\centering\arraybackslash}p{2.4cm} *{7}{>{\centering\arraybackslash}X}}
\toprule
Method & NN & FT & ST & nDCG & E & MRR & mAP \\
\hline
random order & 80.2 & 66.6 & 38.4 & 67.9 & 51.7 & 84.8 & 72.5 \\
w/o local GCN & 80.6 & 67.7 & 39.8 & 68.9 & 53.4 & 85.4 & 74.6 \\
w/o global att & 80.2 & 66.6 & 38.7 & 68.2 & 51.9 & 84.2 & 73.3 \\
w/o view selector& 78.1 & 64.4 & 36.4 & 66.2 & 49.0 & 83.4 & 69.1 \\
random selector & 83.4 & 66.0 & 37.9 & 67.5 & 51.0 & 84.4 & 71.7 \\
\hline
average pooling & 75.8 & 71.9 & 38.8 & 71.2 & 44.9 & 81.3 & 73.7\\
attention pooling & 83.0 & 73.2 & 41.5 & 73.3  & 42.9 & 83.7 & 76.2\\
ViT pooling \cite{zhang2025meha} & 84.2 & 76.5 & 43.2 & 75.6  & 37.8 & 87.3 & 79.8 \\
\hline
1st & 79.6 & 64.9 & 36.8 & 66.6 & 49.6 & 83.7 & 70.1 \\
1st + 2nd & 79.2 & 71.2 & 40.7 & 68.7 & 39.8 & 85.3 & 76.7 \\
\rowcolor{lightpurple}
1st + 2nd + 3rd & \textbf{90.6} & \textbf{76.5} & \textbf{43.2} & \textbf{80.3} & \textbf{35.6} & \textbf{94.3} & \textbf{82.2} \\
\bottomrule
\end{tabularx}
\end{table}

\subsubsection{The Analysis of View Selector for Hierarchical Graph Neural Network}
\begin{figure}[t]
  \centering
  \includegraphics[width=\columnwidth]{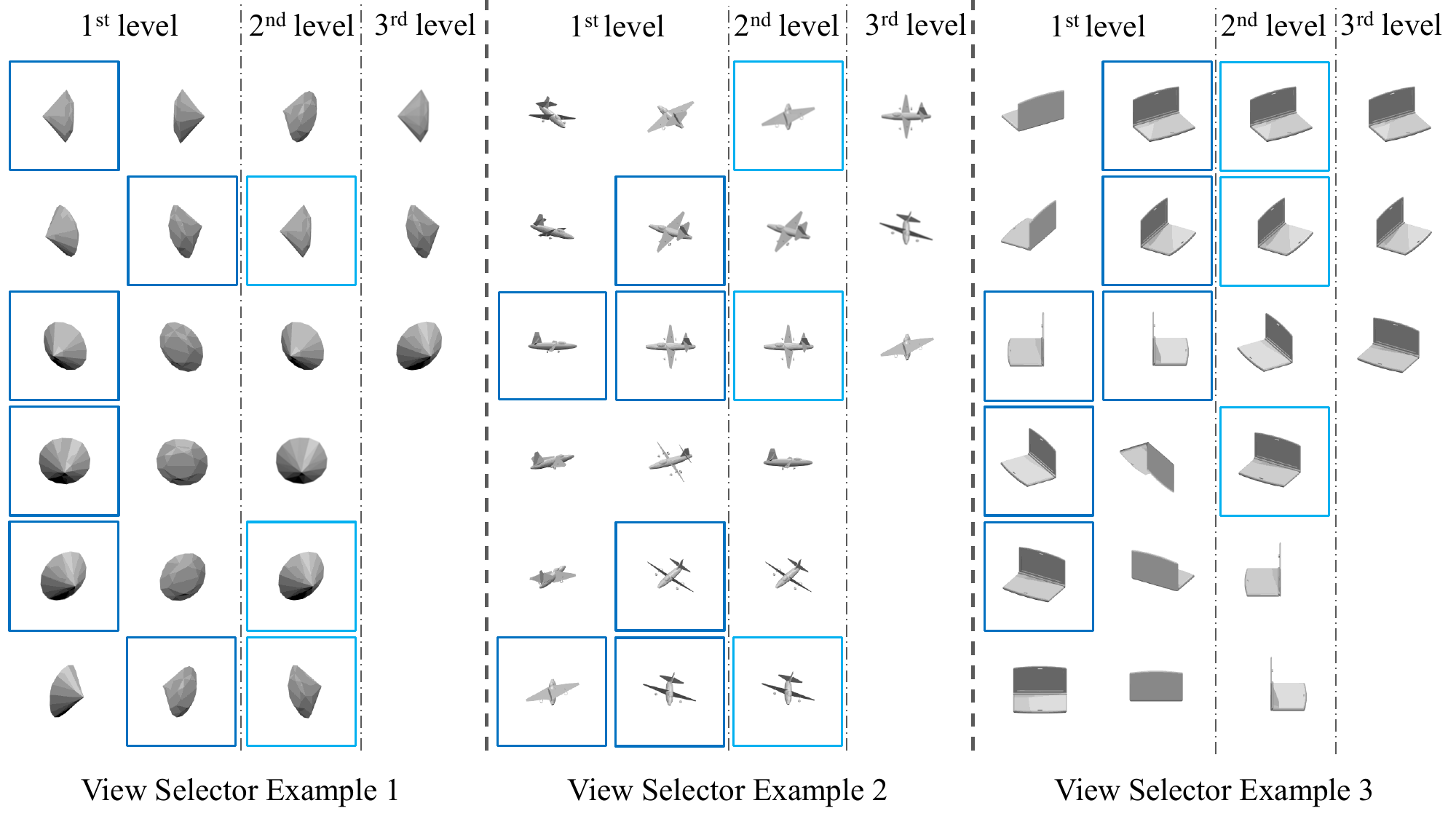}
  \caption{Examples of view selection results on 3D shapes from the SHREC’2014 dataset. Views selected at the first level are highlighted with dark blue boxes, while those selected at the second level are marked with light blue boxes.}
  \label{fig:viewselect}
\end{figure}

To analyze the behavior of the view selector in multi-view encoding, we visualize its scoring distributions and selection patterns. As shown in Fig.~\ref{fig:viewselect}, \textbf{Selector-1} assigns attention scores $\mathbf{F}_{\mathrm{score1}}$ to all candidate views, exhibiting clear spatial discriminability. Views capturing salient geometric regions, prominent silhouettes, or informative poses receive higher scores, while redundant or heavily occluded views are suppressed. The resulting ranking determines the top-selected views (\textit{sel}, marked in red), reflecting a coarse preference for discriminative viewpoints. Based on this reduced set, \textbf{Selector-2} performs a second-stage selection. Its scores $\mathbf{F}_{\mathrm{score2}}$ are more concentrated, indicating a shift from individual view quality to inter-view complementarity. Selector-2 favors combinations that are complementary in pose, silhouette, and local structures, improving geometric coverage while mitigating redundancy-induced overfitting. Together, the two-stage selection produces a more uniform and information-diverse node set for the subsequent GCN, enhancing the expressiveness of camera-space graph convolution.

\subsubsection{The Influence of different Learning Objective for Alignment.}
Table~\ref{tab:Loss_ablation} presents a systematic ablation of our loss design for both category-level and zero-shot SBSR. The results reveal that the two settings rely on distinct objective combinations, validating our task-adaptive formulation. In the category-level setting, all three objectives—classification loss ($\mathcal{L}_{cls}^{ske}$), semantic alignment loss ($\mathcal{L}_{sem}^{ske}$), and quadruplet loss ($\mathcal{L}_{quad}$)—are indispensable. Removing either explicit class supervision from $\mathcal{L}_{cls}^{ske}$ or cross-modal guidance from $\mathcal{L}_{sem}^{ske}$ leads to clear performance degradation (mAP drops to 76.8\% and 74.2\%, respectively), indicating that jointly enforcing classification and semantic alignment is necessary to properly structure the embedding space. Adding $\mathcal{L}_{quad}$ yields a further gain (mAP increases from 77.5\% to 82.2\%), highlighting its role in refining intra-/inter-class boundaries. For zero-shot retrieval, semantic alignment losses for both modalities ($\mathcal{L}_{sem}^{3D}$ and $\mathcal{L}_{sem}^{ske}$) remain fundamental; removing them severely impairs performance, confirming that a shared semantic space is critical for generalization. More importantly, $\mathcal{L}_{quad}$ is the key driver of performance: without it, mAP collapses from 48.5\% to 26.3\%, demonstrating that in the absence of explicit category labels, the fine-grained metric constraints imposed by $\mathcal{L}_{quad}$ are essential for organizing the embedding space and enabling effective retrieval of unseen classes.

\begin{table}[t]
\centering
\caption{Performance on SHREC2013 with different learning objective .}
\label{tab:Loss_ablation}
\setlength{\tabcolsep}{4pt}
\begin{tabularx}{\columnwidth}{p{0.45cm} p{0.45cm} p{0.48cm} *{7}{Y}}

\toprule
\multicolumn{3}{c}{Objective} & \multicolumn{7}{c}{\underline{Category}} \\
\midrule
$\mathcal{L}_{cls}^{ske}$ & $\mathcal{L}_{sem}^{ske}$ & $\mathcal{L}_{quad}$ & NN & FT & ST & nDCG & E & MRR & mAP \\
\midrule
$\times$ & $\checkmark$ & $\checkmark$ & 76.4 & 70.1 & 41.3 & 69.3 & 60.0 & 83.7 & 76.8 \\
$\checkmark$ & $\times$ & $\checkmark$ & 74.2 & 67.1 & 40.2 & 67.6 & 62.1 & 82.3 & 74.2 \\
$\checkmark$ & $\checkmark$ & $\times$ & 78.0 & 71.2 & 41.5 & 69.8 & 61.7 & 82.6 & 77.5 \\
\rowcolor{lightpurple}
$\checkmark$ & $\checkmark$ & $\checkmark$ & \textbf{90.6} & \textbf{76.5} & \textbf{43.2} & \textbf{80.3} & \textbf{35.6} & \textbf{94.3} & \textbf{82.2} \\
\midrule
\multicolumn{3}{c}{Objective} & \multicolumn{7}{c}{\underline{Zero-Shot}} \\
\midrule
$\mathcal{L}_{sem}^{3D}$ & $\mathcal{L}_{sem}^{ske}$ & $\mathcal{L}_{quad}$ & NN & FT & ST & nDCG & E & MRR & mAP \\
\midrule
$\times$ & $\checkmark$ & $\checkmark$ & 28.6 & 23.5 & 22.4 & 23.1 & 65.5 & 48.2 & 32.5 \\
$\checkmark$ & $\times$ & $\checkmark$ & 31.3 & 25.9 & 24.6 & 45.8 & 61.7 & 51.6 & 35.6 \\
$\checkmark$ & $\checkmark$ & $\times$ & 21.7 & 14.2 & 20.3 & 32.5 & 72.1 & 35.4 & 26.3 \\
\rowcolor{lightpurple}
$\checkmark$ & $\checkmark$ & $\checkmark$ & \textbf{42.2} & \textbf{34.3} & \textbf{28.2} & \textbf{52.9} & \textbf{62.5} & \textbf{56.3} & \textbf{48.5} \\
\bottomrule
\end{tabularx}
\end{table}

\subsubsection{The Influence of the Fine-tuning Strategies for Sketch Encoder}
Tab.~\ref{tab:clip_ft} compares fine-tuning strategies for the sketch encoder on SHREC2013. Freezing the ViT-CLIP encoder \cite{Radford2021ICML} yields substantially inferior performance, indicating that task-specific adaptation is necessary for effective sketch representation learning. While full-parameter fine-tuning improves accuracy, it is not the optimal trade-off between adaptation and generalization. In contrast, lightweight adaptation strategies perform better, with Vision Prompt Tuning (VPT) \cite{jia2022vpt} consistently achieving the best results across all metrics. This demonstrates that our use of VPT effectively adapts CLIP to the sketch domain while mitigating catastrophic forgetting, enabling robust alignment with the distinct distribution of hand-drawn sketches.

\begin{table}[t]
  \centering
  \caption{Performance on SHREC2013 with Different Fine-tuning Strategies.}
  \label{tab:clip_ft}
  \begin{tabularx}{\linewidth}{>{\centering\arraybackslash}p{2.8cm}*{7}{>{\centering\arraybackslash}X}}
    \toprule
    Training Strategies & NN & FT & ST & nDCG & E & MRR & mAP \\
    \midrule
    ViT-CLIP frozen & 74.5 & 57.6 & 33.1 & 63.5 & 44.3 & 75.7 & 67.6 \\
    ViT-CLIP all & 85.3 & 70.4 & 38.2 & 70.5 & 51.3 & 93.8 & 74.7 \\
    ViT-CLIP LayerNorm & 87.3 & 73.6 & 42.1 & 73.1 & 53.3 & 94.7 & 79.3 \\
    \rowcolor{lightpurple}
    ViT-CLIP VPT & \textbf{90.6} & \textbf{76.5} & \textbf{43.2} & \textbf{80.3} & \textbf{35.6} & \textbf{94.3} & \textbf{82.2} \\
    \bottomrule
  \end{tabularx}
\end{table}

\subsubsection{The Influence of View Number.}
Tab.~\ref{tab:Number_views} shows that performance improves consistently as the number of views increases. With sparse views, limited angular coverage results in an uneven camera distribution and missing critical geometric cues that cannot be fully recovered by subsequent graph convolutions; these deficiencies propagate through the sparse graph and degrade the overall 3D representation. In contrast, denser and more uniformly distributed views establish a more stable spatial topology, strengthen adjacency structure, and enable richer complementary feature aggregation. Redundant multi-view coverage further mitigates single-view bias and reduces uncertainty, yielding more complete and discriminative 3D shape representations.

\begin{table}[t]
  \centering
  \caption{:Effect of view numbers on SHREC2013.}
  \label{tab:Number_views}
  \begin{tabularx}{\linewidth}{c *{7}{>{\centering\arraybackslash}X}}
    \toprule
    View number & NN & FT & ST & nDCG & E & MRR & mAP \\
    \midrule
    1 view  & 74.5 & 54.1 & 32.2 & 61.6 & 56,7 & 80.7 & 60.4 \\
    3 views & 91.4 & 71.0 & 39.8 & 71.1 & 46.4 & 93.8 & 76.4 \\
    6 views & 92.6 & 73.6 & 42.1 & 73.1 & 43.5 & 94.7 & 80.3 \\
    \rowcolor{lightpurple}
    12 views & \textbf{90.6} & \textbf{76.5} & \textbf{43.2} & \textbf{80.3} & \textbf{35.6} & \textbf{94.3} & \textbf{82.2} \\
    \bottomrule
  \end{tabularx}
\end{table}

\subsubsection{The Influence of Different Training Strategy.}
As shown in Tab. ~\ref{tab:Training_strategy}, the relative effectiveness of one-stage and two-stage training is strongly task-dependent. Two-stage training consistently benefits category-level SBSR, particularly for sketch-to-3D retrieval, by decoupling representation stabilization from semantic alignment under closed-set supervision. In contrast, zero-shot SBSR favors one-stage end-to-end optimization, which better maintains semantic consistency across unseen categories and avoids overfitting to seen-class semantics. This observation highlights an inherent trade-off between discriminability and generalization, motivating our task-specific training design.

\begin{table}[t]
  \centering
  \caption{Performance with different training strategies. 3D to sketch is using the pre-trained sketch encoder and mapping 3D shape into sketch space.}
  \label{tab:Training_strategy}
  \begin{tabularx}{\linewidth}{>{\centering\arraybackslash}p{1.7cm}*{7}{Y}}
    \toprule
    \multicolumn{8}{c}{\underline{Category}} \\
    \midrule
    Method & NN & FT & ST & nDCG & E & MRR & mAP \\
    \midrule
    3D to Sketch & 80.7 & 63.4 & 75.9 & 67.4 & 51.2 & 84.6 & 71.8 \\
    One Stage    & 79.8 & 64.6 & 74.5 & 66.7 & 49.3 & 83.2 & 70.3 \\
    \rowcolor{lightpurple}
    Sketch to 3D & \textbf{93.8} & \textbf{75.5} & \textbf{85.9} & \textbf{74.1} & \textbf{57.7} & \textbf{96.0} & \textbf{82.0} \\
    \midrule
    \multicolumn{8}{c}{\underline{Zero-Shot}} \\
    \midrule
    Method & NN & FT & ST & nDCG & E & MRR & mAP \\
    \midrule
    3D to Sketch & 23.3 & 21.7 & 19.4 & 30.9 & 66.7 & 31.3 & 33.8 \\
    Sketch to 3D & 27.5 & 25.2 & 21.7 & 35.6 & 64.0 & 36.7 & 35.7 \\
    \rowcolor{lightpurple}
    One Stage    & \textbf{42.1} & \textbf{34.3} & \textbf{56.3} & \textbf{52.9} & \textbf{37.5} & \textbf{56.3} & \textbf{48.5} \\
    \bottomrule
  \end{tabularx}
\end{table}

\subsection{Visualization}
\subsubsection{Visualization of Distance Distribution}
\begin{figure}[t]
    \centering
    \begin{minipage}[t]{0.45\linewidth}
        \centering
        \includegraphics[width=\linewidth]{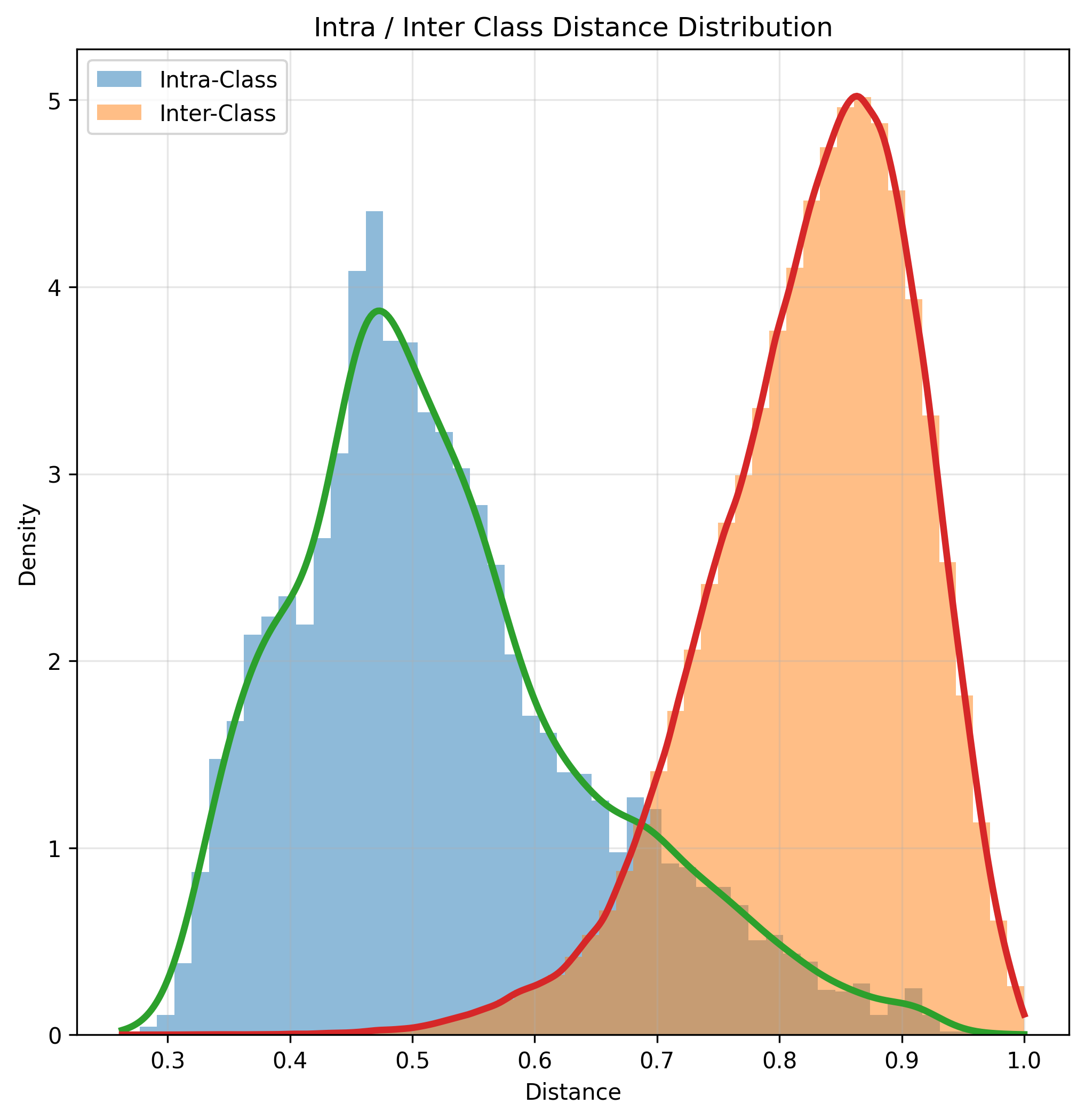}
        \subcaption{Average Pooling(Baseline)}
    \end{minipage}\hfill
    \begin{minipage}[t]{0.45\linewidth}
        \centering
        \includegraphics[width=\linewidth]{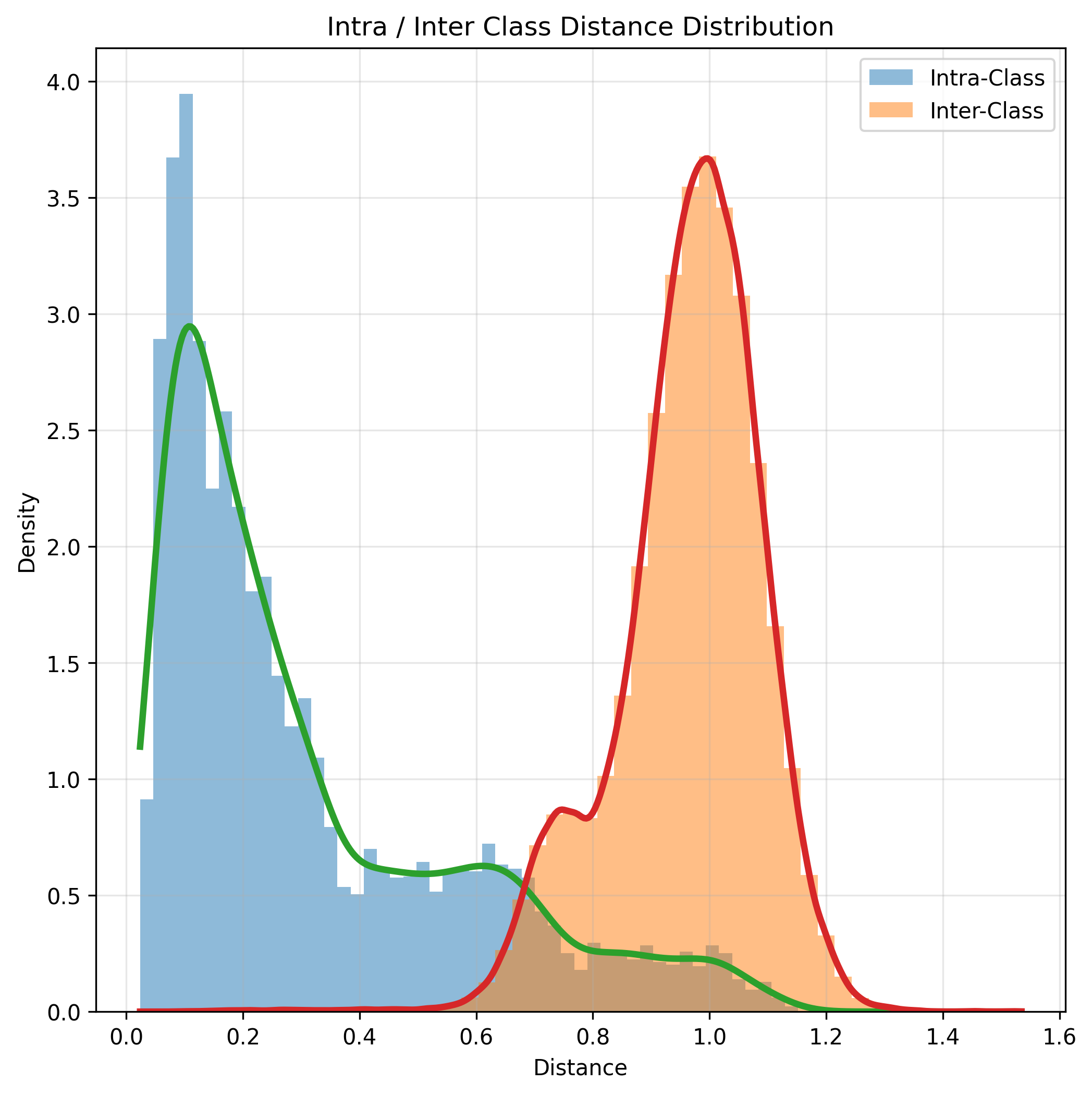}
        \subcaption{Aggregation by MV-HGNN}
    \end{minipage}
    \caption{The frequency of intra-class and inter-class distances of SHREC’2013.}
    \label{fig:distance}
\end{figure}

\begin{figure}[t]
    \centering
    \begin{minipage}[t]{0.45\linewidth}
        \centering
        \includegraphics[width=\linewidth]{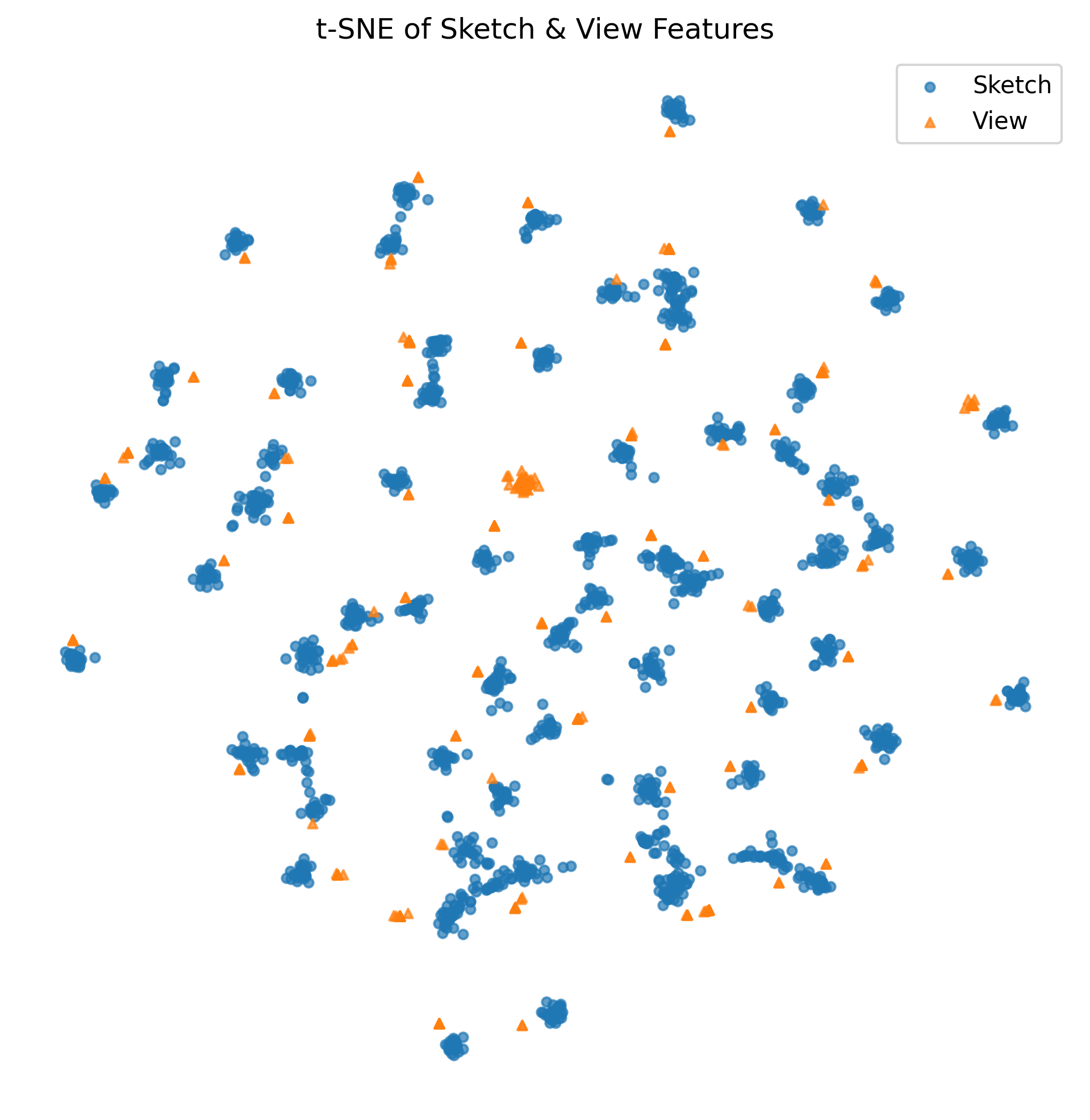}
        \subcaption{Average Pooling(Baseline)}
    \end{minipage}\hfill
    \begin{minipage}[t]{0.45\linewidth}
        \centering
        \includegraphics[width=\linewidth]{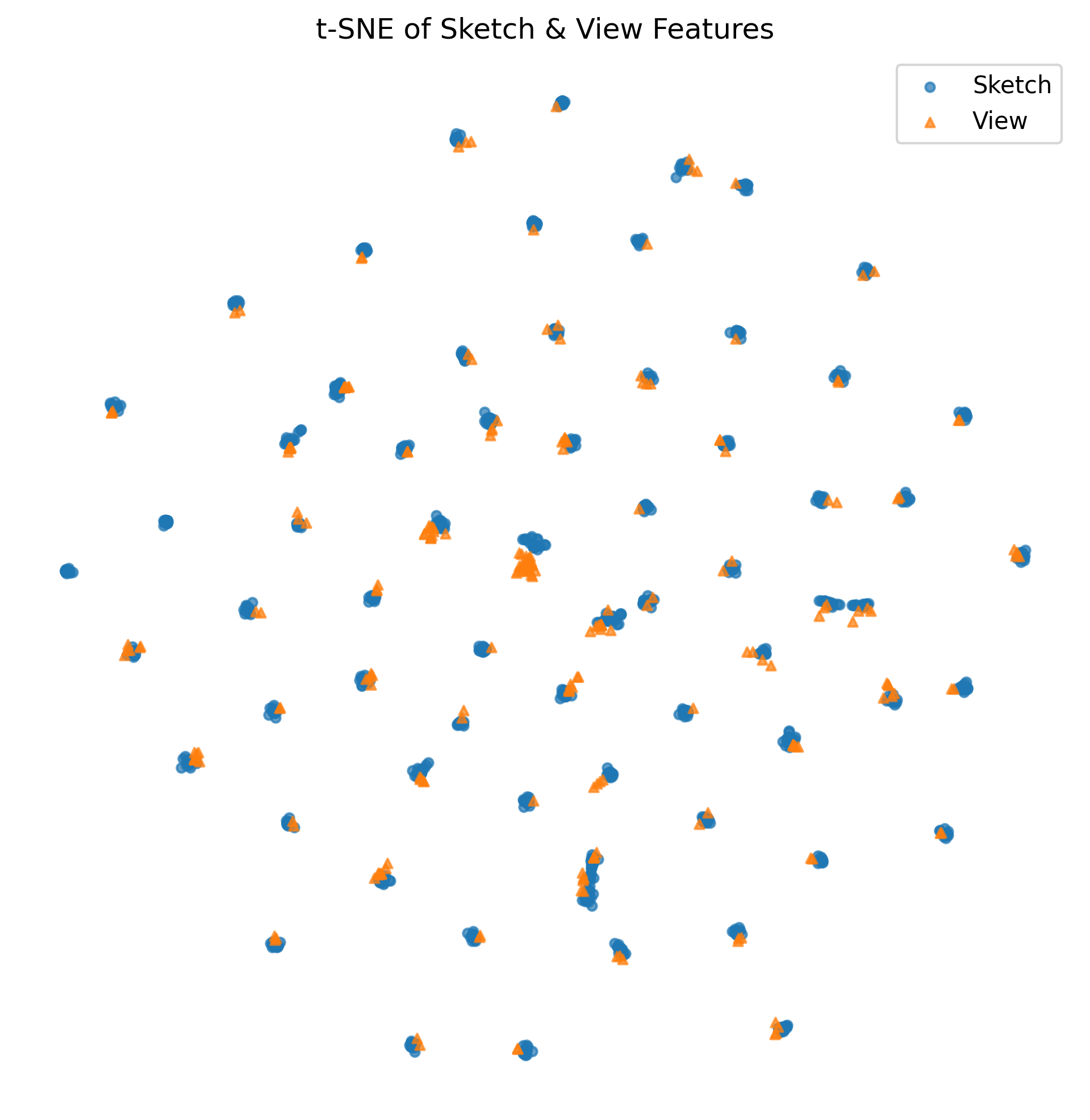}
        \subcaption{Aggregation by MV-HGNN}
    \end{minipage}
    \caption{t-SNE Visualization of test samples on the SHREC’13.}
    \label{fig:kde_tsne}
\end{figure}

To gain insight into the structure of the learned embedding space, we visualize feature distributions using pairwise distance matrices and t-SNE embeddings (Fig.~\ref{fig:kde_tsne} and Fig.~\ref{fig:distance}). As shown in Fig.~\ref{fig:kde_tsne}, our method produces a markedly larger margin between intra-class (blue) and inter-class (orange) distance distributions than the pooling baseline, with substantially reduced overlap. This suggests that our representations form more compact and well-separated semantic clusters. Fig.~\ref{fig:distance} further supports this finding through a direct t-SNE comparison. Our embeddings exhibit clearly cohesive and distinct clusters, whereas the baseline yields dispersed and overlapping patterns, indicating weaker semantic organization. Together, these visualizations highlight a key advantage of our hierarchical aggregation: by explicitly preserving cross-view geometric structure that is discarded by naive pooling, our approach learns representations with significantly stronger discriminative power.

\subsubsection{Visualization of Retrieval Examples}
\begin{figure}[t]
  \centering
  \includegraphics[
    width=\linewidth,
    trim=0mm 0mm 0mm 10mm,
    clip
]{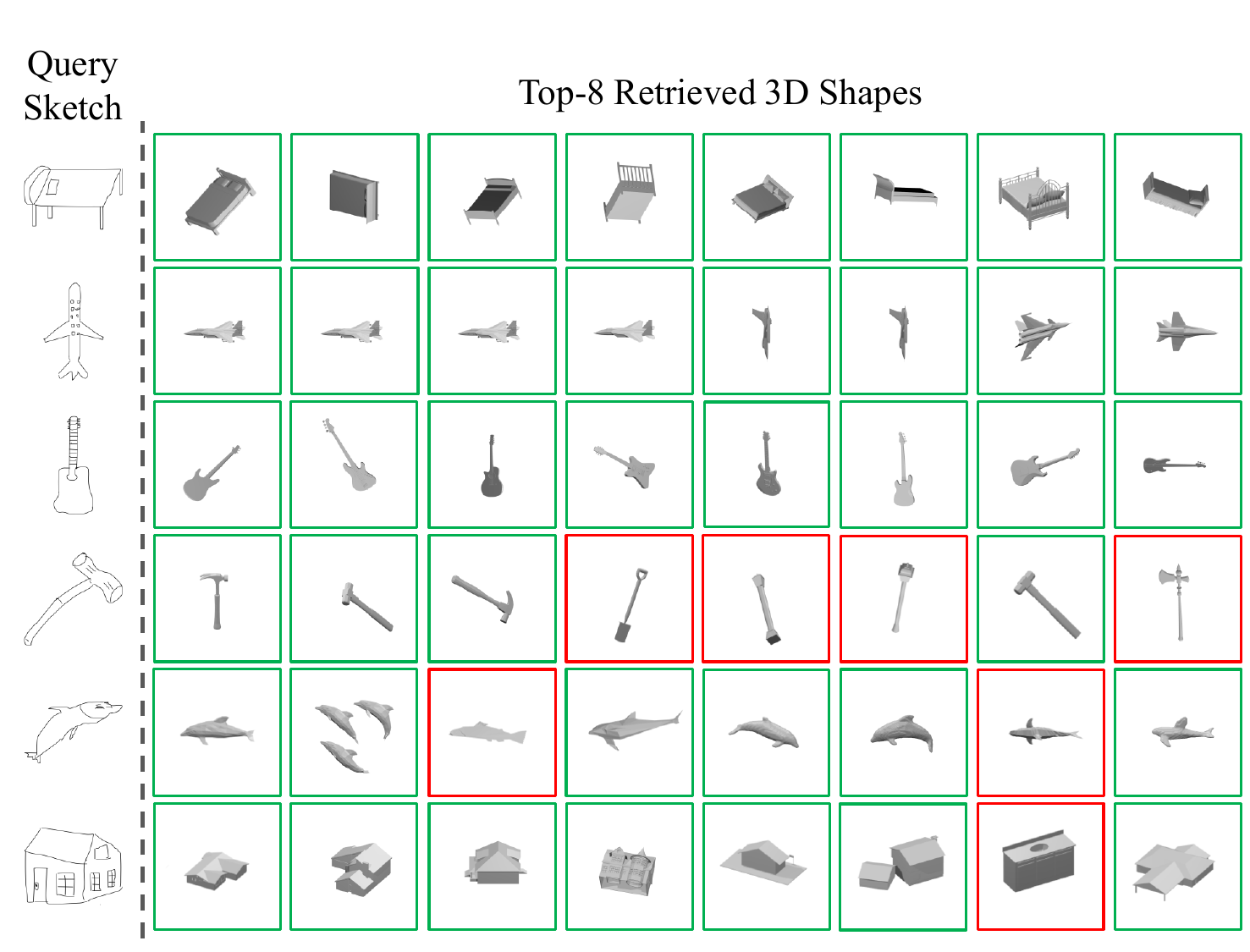}
  \caption{Retrieval examples for sketch-based 3D shape retrieval on SHREC’2014.
Correctly retrieved shapes are highlighted in green, whereas incorrect candidates are in red.}
  \label{fig:Results}
\end{figure}

Fig.~\ref{fig:Results} shows representative retrieval results on the SHREC’2013 dataset. For each query sketch (left), we report the top-8 retrieved 3D shapes. Our method effectively captures semantic correspondences, enabling accurate retrieval of relevant shapes in most cases. The top rows illustrate correct matches even under substantial shape variations and local geometric discrepancies. Occasional errors arise due to the limited number of correct matches, such as fish being confused with dolphins, reflecting their visual similarity; these cases are rare and do not significantly impact overall performance.

\section{Conclusion}
In this work, we address sketch-based 3D shape retrieval by identifying weak multi-view feature aggregation as a key bottleneck for both retrieval performance and generalization. To tackle this, we propose a Multi-View Hierarchical Graph Neural Network (MV-HGNN) that constructs a view-level graph and progressively aggregates multi-view features via local graph convolution and global attention, with a learnable view selector to reduce redundant views. This hierarchical representation preserves local geometric consistency and global structural information, enabling more effective cross-modal alignment. We further introduce a prototype-based alignment strategy supported by CLIP text embeddings, projecting sketches and 3D shapes into a shared semantic space. This allows the same model to handle both category-level and zero-shot retrieval, adapting to more general scenarios. Extensive experiments on public benchmarks show that MV-HGNN consistently outperforms state-of-the-art methods. In future work, we plan to explore fine-grained SBSR and develop more general paradigms applicable to broader cross-modal and 3D retrieval scenarios.

\begin{acks}
This research was supported by external funding.
\end{acks}

\bibliographystyle{ACM-Reference-Format}
\bibliography{sample-base}

\end{document}